\documentclass[11pt]{article}
\usepackage[preprint]{acl}
\usepackage{times}
\usepackage{latexsym}
\usepackage[T1]{fontenc}
\usepackage[utf8]{inputenc}
\usepackage{microtype}
\usepackage{graphicx}
\usepackage{booktabs}
\usepackage{tabularx}
\usepackage{amsmath,amssymb}
\usepackage{algorithm}
\usepackage{algpseudocode}
\usepackage{hyperref}
\hypersetup{pdfauthor={Aolin Li; Xiao Xu; Haihua Xie},pdftitle={SceneWeave: Audience-Guided Scene Exploration for Dramatic Adaptation},pdfsubject={},pdfkeywords={dramatic adaptation, screenplay generation, scene exploration}}

\title{SceneWeave: Audience-Guided Scene Exploration for Dramatic Adaptation}
\author{Aolin Li$^{1,2}$ \quad Xiao Xu$^{3}$ \quad Haihua Xie$^{4,\ast}$\\
\textnormal{\small $^{1}$Department of Science Theatre, The Central Academy of Drama, Beijing, China}\\
\textnormal{\small $^{2}$Hithink RoyalFlush Information Network Co., Ltd., Hangzhou, China}\\
\textnormal{\small $^{3}$China Telecom Artificial Intelligence Technology (Beijing) Co., Ltd., Beijing, China}\\
\textnormal{\small $^{4}$Beijing Institute of Mathematical Sciences and Applications (BIMSA), Beijing, China}\\
\textnormal{\small $^{\ast}$Corresponding author: haihua.xie@bimsa.cn}}

\begin{document}
\maketitle

\begin{abstract}
Adapting prose into a script requires transforming narration and inner experience into scenes that convey relationships and psychological contradictions. Planning specifies intended effects but leaves their realization in dialogue and action unresolved. Drawing on screenwriters' accounts of revising scenes and their connections to the surrounding work, we present SceneWeave, which uses alternative scene drafts to revise the adaptation as a whole. It writes contrasting treatments of a focal scene, compares the interactions they create, and revises characterization, preceding preparation, subsequent development, and scene order before generating the complete script. The central contribution is to use actual scene writing to inform adaptation decisions and propagate their consequences across the work. We also develop a task-specific writing model and a human-preference model for comparing script drafts. The writing model learns from adaptation examples and human preferences; the preference model combines human and synthetic supervision. Blind reader comparisons favor SceneWeave over a baseline workflow and description-based controls, while content assessment indicates reduced narrative redundancy and strong psychological treatment.

\end{abstract}

\section{Introduction}

Prose conveys relationships through narration, perspective, memory, and unspoken thoughts. Adaptation must decide how these meanings emerge in performance. It may bring separated characters together, redistribute initiative, replace an event, or let an ordinary exchange carry a psychological contradiction. The resulting script determines whether these changes work.

\begin{figure*}[t]
    \centering
    \includegraphics[width=\textwidth]{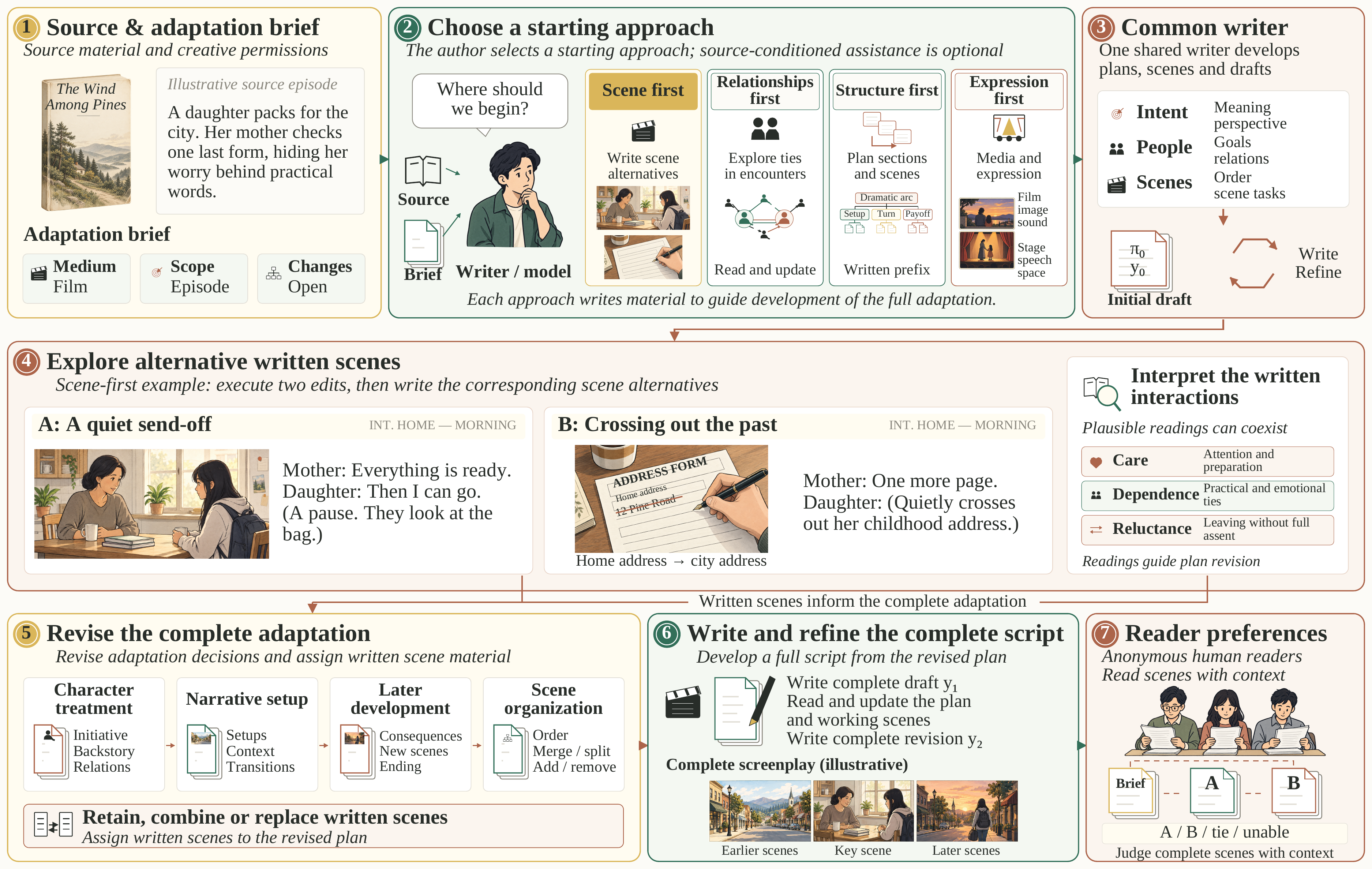}
    \caption{SceneWeave. The scene-first example develops two treatments of a mother--daughter encounter and uses interpretations of those scenes to revise the adaptation (4--5). All four approaches provide material for complete writing (6); anonymous reader preferences evaluate the resulting scenes (7). INT. denotes an interior location in the screenplay headings. $\pi_0$ is the initial plan; $y_0,y_1,y_2$ denote the initial, developed, and revised drafts.}
    \label{fig:framework}
\end{figure*}

Large language model (LLM) writing systems support detailed planning, candidate selection, and local and global revision \citep{yang-etal-2022-re3,yang-etal-2023-doc,wu-etal-2026-superwriter,xie-etal-2025-dramaturge}. Novel-to-screenplay generation has also been investigated \citep{lin-etal-2025-reader-rewriter}. We examine whether writing alternative scenes before finalizing an adaptation improves decisions about characterization and scene organization.

In Gerwig's adaptation of \textit{Little Women}, an outline could reduce Amy's wish to marry a wealthy man to a calculating trait. In the film, Laurie appeals to love, and Amy counters that marriage determines her financial and legal future; Gerwig reports writing much of this exchange shortly before filming \citep{gerwig-august-2020-scriptnotes}. Their disagreement brings desire, dependence, and unequal freedom into the same encounter. If Amy simply announced her plan, the choice would carry a different meaning. Writing both treatments would expose that difference and help revise her earlier characterization and the later development of their relationship.

Screenwriting practice motivates revisiting the connections between a scene and the surrounding work. Sciamma describes a late scene invention that links a love story with its later memory \citep{sciamma-2019-lecture}; Gerwig discusses how interleaving timelines change the emotional meaning of encounters \citep{gerwig-august-2020-scriptnotes}. Birch describes discovering possibilities through formal constraints, while Straughan relates public ritual to private conduct in a particular adaptation \citep{birch-2024-lecture,focus-2024-straughan}. These accounts inform our treatment of expression and arrangement as revisable creative decisions.

SceneWeave uses alternative scene drafts to revise the adaptation plan and generate the complete script (Figure~\ref{fig:framework}). We evaluate the full method against a planning-and-refinement baseline and examine the contribution of written scenes through two mechanism controls. The implementation is available on GitHub\footnote{\url{https://github.com/Aolin-MIR/SceneWeave}}, with model and evaluation artifacts in a companion Zenodo record\footnote{\url{https://doi.org/10.5281/zenodo.23329962}}.

\section{Related Work}
\label{sec:related-work}

\paragraph{Planning and revising narratives.}
Recursive Reprompting and Revision (Re3) combines an overarching plan, recursive contextual prompting, reranking, and consistency editing \citep{yang-etal-2022-re3}; Detailed Outline Control (DOC) uses detailed hierarchical outlines and generation control \citep{yang-etal-2023-doc}. Agents' Room separates planning and writing through a shared scratchpad \citep{huot-etal-2025-agents-room}. SuperWriter combines planning, incremental writing, and refinement with stage-specific supervision and hierarchical direct preference optimization (DPO) \citep{wu-etal-2026-superwriter}.

\paragraph{Screenplay generation and reflection.}
Dramatron hierarchically generates characters, plot beats, locations, and dialogue for human co-writing \citep{mirowski-etal-2023-dramatron}. Reader--Rewriter ($R^2$) extracts novel characters and event graphs, then creates outlines and scenes with hallucination-aware refinement \citep{lin-etal-2025-reader-rewriter}; its graph traversals support main-storyline, chronological, and chapter-order adaptation. Dramaturge coordinates global, scene-level, and cross-scene revisions \citep{xie-etal-2025-dramaturge}, while DuoDrama generates inner experience and actor-perspective questions for human reflection \citep{tang-etal-2026-duodrama}.

\paragraph{Preference learning for writing.}
Language Model Authored, Manually Polished (LAMP) provides professional paragraph edits and a taxonomy of recurring problems \citep{chakrabarty-etal-2025-lamp}. Writing Quality Reward Models learn from edits and select paragraph revisions \citep{chakrabarty-etal-2025-writing-rewards}; LitBench trains verifiers on human story preferences and evaluates new story rankings \citep{fein-etal-2026-litbench}. Our assessment of presentation-order reliability is motivated by \citet{wang-etal-2024-fair-evaluators}.

\section{Task and Preference Objective}

\subsection{Adaptation permissions}

Given source prose $x$ and adaptation brief $b$, the task produces a complete script $y$ at the requested scope. The brief specifies medium, fixed requirements, and permissions over eight edit axes: plot, character actions, relationships, ending, background, props, scene organization, and expression. The study uses prose passages to produce self-contained short dramatic works.

A structured adaptation plan is $\pi=(\eta,c,S)$, with ordered scenes $S=(s_1,\ldots,s_m)$, where $\eta$ describes the interpretation, $c$ records choices on the edit axes, and each scene $s_i$ has a unique identifier and a textual description of its intended development. These decisions remain revisable during scene development.

\subsection{Dramatic expression}

Readers judge how the permitted adaptation expresses relationships, inner contradictions, and social circumstances through speech, action, silence, and scene arrangement. A reply may sustain a relationship while withholding assent; setup and aftermath may change the meaning of an encounter.

For complete candidates $y_a,y_b$, the comparator predicts $\mathbf p_\phi=P_\phi(r\mid x,b,y_a,y_b)$ over $r\in\{A,B,\mathrm{tie}\}$, using their common source and creative brief. Human evaluation also records when readers are unable to judge because the supplied material is insufficient.

\section{Scene Exploration and Complete Writing}
\label{sec:scene}

\begin{figure*}[t]
    \centering
    \includegraphics[width=\textwidth]{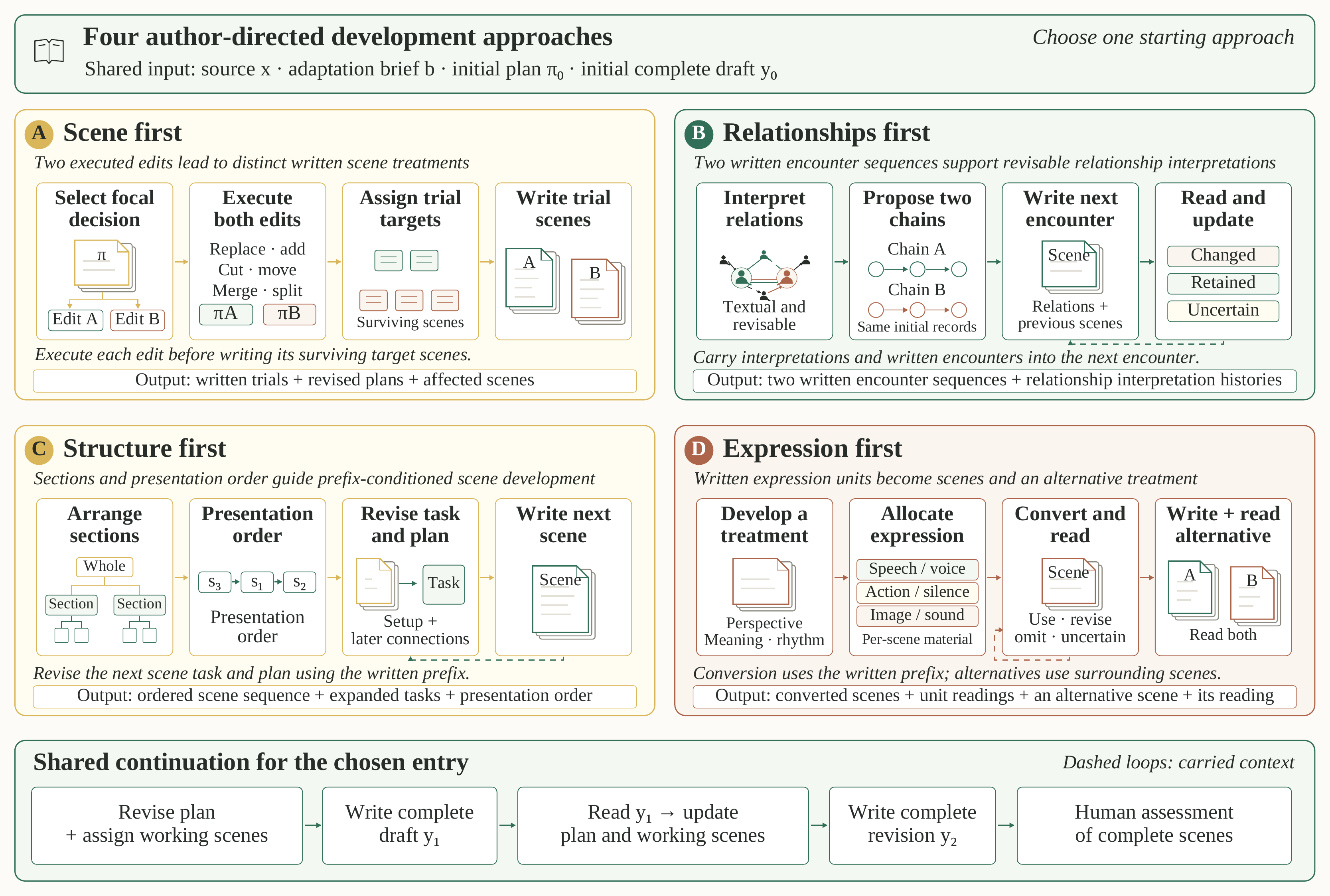}
    \caption{Four adaptation strategies. Solid arrows indicate generation order; dashed arrows supply interpretations or written context to later scenes. Each strategy contributes material for plan revision and complete-script generation.}
    \label{fig:approaches}
\end{figure*}

\subsection{Initial plan and alternative edits}

Matched conditions share an initial plan $\pi_0$ and complete draft $y_0$, generated with the released SuperWriter planning and writing prompts \citep{wu-etal-2026-superwriter} and fixed before method-specific development. The controlled study uses the common API model for all functions.

The model takes $x,b,\pi_0,y_0$ as input and produces a provisional plan $\pi$, identifier $k$ of the scene to explore, links between affected scenes $D$, and two edits $o_0,o_1$. Each edit specifies changes to $\pi$, their intended benefit, and potential drawbacks. Applying edit $o_j$ gives $\pi_j=\operatorname{Apply}(\pi,o_j)$, for $j\in\{0,1\}$.
The resulting plans differ from the original and from each other while exploring the same focal decision. Changes may affect participants, actions, circumstances, or expression within the brief's permissions; the edited plans determine the scenes to be written.
Trial writing follows the edited scene treatment within the brief's fixed requirements. If the edit conflicts with earlier choices for the whole adaptation, those choices remain provisional until the plan is revised using both trials.

Edits replace, insert, remove, move, merge, or split scenes, and revise the intended interpretation or a choice along an edit axis. Each edit is applied to the plan before writing the scenes that remain (Appendix~\ref{app:implementation}).

The links $D$ identify related scenes whose characterization, setup, aftermath, or position may need revision after the focal edit.

\subsection{Writing the exploratory scenes}

After each alternative edit is applied, the surviving scenes selected for trial writing are ordered as they appear in the revised plan: $K_j=(k_{j1},\ldots,k_{jr_j})$. These can include several scenes after a merge, split, or change to neighboring encounters. The shared writing model generates the corresponding scene texts $v_j=(v_{j1},\ldots,v_{jr_j})$ in that order. Each scene $v_{j\ell}$ is conditioned on $h_j=(x,b,\pi_0,y_0,\pi_j,o_j,D)$, its target identifier $k_{j\ell}$, and the previously written scenes $v_{j,<\ell}$.
Each output contains a scene identifier and complete dramatic text. Earlier scenes from the same alternative provide writing context.

\subsection{Revising characters and the surrounding work}

Given the source, initial plan and draft, provisional plan, and both scene alternatives, the model proposes plan edits, revisions to surrounding scenes, and assignments of draft material:
\begin{equation}
 \begin{aligned}
 (o',\delta,W)&\sim G_{\mathrm{update}}(\cdot\mid x,b,\pi_0,y_0,\pi,v_0,v_1,D),\\
 \pi'&=\operatorname{Apply}(\pi,o').
 \end{aligned}
 \label{eq:update}
\end{equation}
$\delta$ records the original and revised treatments, affected scenes, and whether characterization, setup, aftermath, and scene order change. For instance, a revised encounter may change the meaning of an earlier gift, the response that follows, or the attention distributed across neighboring scenes.

Each assignment $w\in W$ associates a scene in $\pi'$ with stored trial scenes, other contributing scenes, a rewriting focus, and potential drawbacks. Assignments follow the revised scene order. They can combine either alternative with the common draft, redistribute interaction across scenes, or specify fresh writing. The assigned text is supplied for rewriting within the revised arrangement.

\subsection{Complete writing, review, and revision}

The writing model receives the source, brief, revised plan, shared initial plan and draft, decisions about surrounding scenes, and designated material, producing $y_1$. The model then reviews $y_1$ alongside the initial draft and both scene alternatives. It considers how shifts between speech shared by characters, private thought, and speech or voice addressed to the audience alter who can respond and which responses and actions become possible in the next interaction.

The review produces revision proposals $q$. Given the same context, the model decides which proposals to adopt and updates the plan and scene assignments. The writing model implements the highest-priority accepted proposal using the revised context and $q$, producing $y_2$.

Algorithm~\ref{alg:scene} summarizes the procedure. A final comparison checks unintended repetition; the criterion for retaining the previous draft is given in Appendix~\ref{app:implementation}.

\begin{algorithm}[t]
\caption{Scene-first adaptation}
\label{alg:scene}
\begin{algorithmic}[1]
\Require Source $x$, brief $b$, common $(\pi_0,y_0)$
\State Propose $(\pi,k,D,o_0,o_1)$
\For{$j\in\{0,1\}$}
 \State Set $\pi_j\gets\operatorname{Apply}(\pi,o_j)$; derive $K_j$
 \State Set $h_j\gets(x,b,\pi_0,y_0,\pi_j,o_j,D)$ and $v_j\gets()$
 \For{$\ell=1,\ldots,|K_j|$}
  \State Write $v_{j\ell}$ from $h_j,k_{j\ell},v_j$; append to $v_j$
 \EndFor
\EndFor
\State Read both trials; propose $(o',\delta,W)$
\State Apply $o'$; supply the written scenes assigned by $W$
\State Write complete draft $y_1$ from $\pi'$ and assigned material
\State Review $y_1$; revise $\pi'$ and scene assignments
\State Write complete revision $y_2$ from the revised plan
\If{$y_2$ meets the unintended-repetition criterion}
 \State \Return $y_1$
\Else
 \State \Return $y_2$
\EndIf
\end{algorithmic}
\end{algorithm}

\section{Preference Learning and Adaptation Strategies}

\subsection{Learning audience preferences}
\label{sec:evaluator}

We fine-tune a two-billion-parameter decision model as the comparator using LoRA \citep{hu-etal-2022-lora}, retaining its public pointer head for A, B, and tie \citep{strands-2026-decider}. Its 3,000 training comparisons comprise 269 human and 2,731 synthetic records; six human records express indifference. Human labels come from Arena, PPE, and HelpSteer2/3 \citep{arena-public-data,frick-etal-2024-ppe,wang-etal-2024-helpsteer2-preference,wang-etal-2025-helpsteer3-preference}. Screening retains dramatic-writing requests with their context and original labels. Human examples guide synthetic annotation. Study partitions group related works, tasks, authors, and adaptations (Appendix~\ref{app:data}).

Let $\mathbf q_i=(q_{iA},q_{iB},q_{iT})$ contain the observed human vote fractions. When human votes are divided between A and B, their observed fractions are retained; explicit indifference contributes to $q_{iT}$. For synthetic non-tie preferences, $t$ denotes the probability of preferring A, and the target distribution is $(0.8t+0.1,\,0.8(1-t)+0.1,\,0)$.

Pairs are presented in orders A--B (AB) and B--A (BA). With comparison weight $w_i$ and $P$ exchanging the A/B coordinates while retaining tie, bidirectional training minimizes
\begin{equation}
 \begin{aligned}
 \mathcal L_{\mathrm{pref}}&=-\frac{1}{2\sum_i w_i}\sum_i w_i\bigl[\mathbf q_i^\top\log\mathbf p_i^{AB}\\
 &\qquad +(P\mathbf q_i)^\top\log\mathbf p_i^{BA}\bigr].
 \end{aligned}
 \label{eq:preference-loss}
\end{equation}
During training, the positions of the A, B, and tie response options are permuted independently of the preference labels; pointer indices are aligned to their semantic classes. Reversing candidate presentation swaps the A/B target coordinates. Human preference judgments in the development set are used to assess agreement, calibration, and coverage across scenes, connected scenes, continuations, and complete scripts.

\subsection{Alternative adaptation strategies}

Authors choose an approach according to their intended treatment. An optional LLM recommendation uses the source and brief to compare the opportunities and risks of the four strategies. Strategy order is randomized; writing receives the chosen strategy and the shared initial plan and draft.

\paragraph{Relationships.}
This approach writes two alternative sequences of linked encounters to develop particular relationships. Each encounter is written from the current interpretation and preceding encounters. After each encounter, a separate interpretation step revises the account of how characters act toward one another, retaining unresolved questions and continuity with earlier encounters; the revised interpretation informs the next encounter and plan revision. The character-interpretation step uses an adapted version of Dramaturge's publicly released instruction \citep{xie-etal-2025-dramaturge}.

\paragraph{Structure.}
The provisional scenes are grouped into sections with an explicit presentation order, including nonlinear arrangements. Before writing each scene, its task is expanded using the written prefix and current arrangement, and the plan is revised accordingly. This procedure builds on hierarchical outline planning \citep{yang-etal-2023-doc,wu-etal-2026-superwriter}.

\paragraph{Expression.}
A scene-level treatment describes what narrative perspective, imagery, rhythm, or interior expression contributes. The approach assigns expressive units---speech, voice, action, silence, image, sound, space, and time---and converts them into scenes using the written prefix. Interpretation records indicate how these units are used, transformed, or omitted, and where their treatment remains uncertain. Once the sequence is written, the generator proposes a different arrangement of expressive elements and writes an alternative scene in its preceding and following context. An interpretation step compares the two scene treatments without favoring any expressive channel. The original sequence and alternative scene provide separate inputs for updating the full adaptation plan.

These approaches feed into the plan-update and complete-writing sequence in Section~\ref{sec:scene}, using their scene drafts or encounter sequences (Figure~\ref{fig:approaches}).

\section{Evaluation}

\subsection{Sources and reproducible conditions}

Evaluation draws on a privately held prose collection. Twenty prose-fiction works from 17 author groups are held out using a deterministic ordering of work identifiers within a work-level partition. Each eligible work contributes a supplied source passage, preceding prose context, and adaptation brief. Source paragraphs are extracted directly from the prose, without using screenplays to retrieve the source passages. Known work, adaptation-family, and author links and actual text fingerprints are checked against study training inputs and targets. Passage reviews specify their bounded scope. Source manifests retain medium, language, scope, fixed requirements, and all permitted edits.

All four conditions use \texttt{gpt-6-luna} for writing, planning, plan revision, and script review. They share the source, brief, initial plan and draft, and instructions for common tasks. Prompts, decoding settings, and source selection are fixed before evaluation.

Coverage and text diagnostics are reported by work and condition.

\subsection{Comparisons and mechanism controls}

\paragraph{Detailed scene planning.}
This control uses the same proposed edits, target scenes, and intended treatments as SceneWeave. It replaces each written scene trial with detailed scene planning that can describe exchanges, psychological possibilities, imagery, rhythm, setup in preceding scenes, and connections to later scenes. The control describes possible spoken exchanges without writing the final dialogue. The descriptions replace written trials during plan revision and complete-script generation. This contrast tests the combined contribution of written scene material to plan revision and complete writing.

\paragraph{SuperWriter prompt workflow.}
The baseline uses the official SuperWriter planning, writing, and refinement prompts \citep{wu-etal-2026-superwriter}. After the common initial draft, it performs refinement, full-script review, and revision with the same LLM as SceneWeave. The complete published prompts are preserved; only output-field names are adapted to the common script representation. All four conditions use the same final repetition check.

\paragraph{Written scenes versus descriptions for plan revision.}
This control holds the source, initial draft, plan, edits, and written scene alternatives fixed. Plan revision receives either the scenes or descriptions of their intended treatments. Subsequent writing and review have access to the same scene texts, selected according to the revised plan. The contrast isolates the contribution of written scenes to the first plan-revision decision. The controlled study focuses on the scene-first strategy.

\subsection{Outcome evaluation}
\label{sec:human}

The primary outcome is blind reader preference on complete scenes with their preceding and following passages. Before generation or scoring, each work receives a source situation anchor. Scenes are matched by roles and situations under permitted changes to cast, setting, actions, and ending; unreliable correspondences are unassessable. Readers receive the source facts and common brief, start with one pair, and may continue with other works. Assignment balances completed judgments across works and contrasts. The judgment covers the displayed encounter and its context.

Close reading examines changes in scene arrangement, dialogue, action, and relationships relative to the shared initial draft. Content assessment summarizes these properties across complete scripts.

Readers choose A, B, tie, or unable to judge and provide a brief reason grounded in the scene and its supplied context. Method identities and model judgments are hidden. Appendix~\ref{app:human} details presentation and assignment.

Human preference scores are 1 for SceneWeave, 0 for the baseline, and 0.5 for a tie. Unable-to-judge responses count toward coverage and are excluded from the score. Scores are averaged within works, then across works. SuperWriter is the primary contrast; the two controls are secondary.

\begin{table*}[!t]
\centering
\begingroup
\setlength{\tabcolsep}{10pt}
\renewcommand{\arraystretch}{1.15}
\begin{tabularx}{\textwidth}{@{}Xccc@{}}
\toprule
Comparison & Wins / Ties / Losses & Preference (\%) & Works\\
\midrule
SceneWeave vs. SuperWriter prompts & 11 / 1 / 4 & 71.9 & 4\\
Scene exploration vs. detailed planning & 5 / 0 / 3 & 62.5 & 2\\
Written vs. described input for plan revision & 4 / 3 / 1 & 68.8 & 2\\
\bottomrule
\end{tabularx}
\endgroup
\caption{Blind scene preferences from four theatre graduate students. Wins, ties, and losses count individual judgments; preference is the mean of work-level scores, with ties assigned 0.5. The Works column gives the number of distinct source works, each judged by all four readers.}
\label{tab:preferences}
\end{table*}

\section{Results}

\subsection{Human scene preferences}

Four theatre graduate students each judged eight anonymous bilingual pairs from distinct works: four against SuperWriter and two for each mechanism control, yielding 32 judgments. English originals were displayed alongside Chinese translations. Pairs were selected to keep the reading load manageable while preserving complete scenes, without consulting preference scores. All conditions used the common API model. Four judgments were ties; none were abstentions.

Table~\ref{tab:preferences} shows more favorable judgments for SceneWeave in all three contrasts. Against the SuperWriter prompt workflow, three works favored SceneWeave and one was evenly divided; excluding ties, its win rate was 11/15 (73.3\%). The detailed-planning contrast split by work, while one plan-revision comparison received three ties. Mean pairwise agreement among readers was 50\% against SuperWriter, 75\% for detailed planning, and 50\% for the plan-revision contrast; agreement is the proportion of reader pairs making the same A, B, or tie choice within each work.

\subsection{Complete-script content diagnostics}
\label{sec:auxiliary-results}

The complete-script comparison uses fifteen works with complete drafts and candidate-selection judgments for every method, excluding two method-development works. Cohort membership is fixed by availability before outcome analysis. Methods share the source, brief, initial plan, and initial script. SceneWeave generates one candidate per adaptation strategy; each baseline receives four generation opportunities. Within each method, the frozen preference model compares these candidates with the initial script and selects the candidate with the most wins minus losses, using fixed tie-breaking. The selected scripts, including retained initial drafts, enter the content assessment below.

The baselines are the released SuperWriter prompt workflow, a component-level implementation of Dramaturge \citep{xie-etal-2025-dramaturge}, and a strong prompt workflow using approach-specific writing instructions. Dramaturge retains global and scene-level review, separate description and dialogue editing, and structural and detail revision phases. Appendix~\ref{app:implementation} gives its stopping conditions. All methods use \texttt{gpt-6-luna}. 

We adapt the five DramaBench content dimensions to complete adaptations \citep{ma-etal-2026-drama-failures}. Every method is assessed on the same fifteen works using the original brief. \texttt{gpt-6-luna} annotates the frozen selected scripts with method identities concealed; identical scripts share annotations. Deterministic units delimit action paragraphs, complete dialogue turns, and scene boundaries.

Narrative redundancy, character inconsistency, distinctive dialogue and world-continuity errors are calculated from categorical labels. Psychological and conflict treatment use three separately anchored criteria each. Psychological treatment assesses particularity, realization and development; conflict assesses tension, development and integration within the brief. Levels are zero for a substantive failure, one for legible but generic or thin treatment, two for convincing specific treatment, and three for finely differentiated and integrated treatment. The grade for each reading is the median of the three criterion levels; the grades from two readings are averaged. 

\begin{table*}[t]
\centering\small
\begingroup
\renewcommand{\arraystretch}{1.1}
\setlength{\tabcolsep}{5pt}
\begin{tabularx}{\textwidth}{@{}Xcccccc@{}}
\toprule
 & Narrative & \multicolumn{2}{c}{Character} & Emotion & Logic & Conflict\\
\cmidrule(lr){2-2}\cmidrule(lr){3-4}\cmidrule(lr){5-5}\cmidrule(lr){6-6}\cmidrule(lr){7-7}
Method & Redundancy (\%)$\downarrow$ & OOC (\%)$\downarrow$ & Voice (\%)$\uparrow$ & Grade$\uparrow$ & Errors (\%)$\downarrow$ & Grade$\uparrow$\\
\midrule
SceneWeave & 0.05 & 0.00 & 69.64 & 2.90 & 1.33 & 2.60\\
SuperWriter prompts & 0.09 & 0.00 & 71.10 & 2.87 & 1.33 & 2.67\\
Dramaturge workflow & 2.77 & 0.00 & 65.28 & 2.73 & 1.33 & 2.57\\
Strong prompt & 0.06 & 0.00 & 66.63 & 2.90 & 2.44 & 2.67\\
\bottomrule
\end{tabularx}
\endgroup
\caption{Complete-script content assessment on fifteen matched works. Redundancy, out-of-character dialogue (OOC), and world-continuity errors are lower-is-better rates; character-specific voice is higher-is-better. Rates are percentages averaged across scripts. Emotion and Conflict use higher-is-better anchored grades (0--3), combining three criteria and two readings.}
\label{tab:dramabench-items}
\end{table*}

Table~\ref{tab:dramabench-items} shows the lowest mean redundancy for SceneWeave and the same mean emotion grade as the strong prompt workflow. SuperWriter has the highest distinctive-dialogue rate; SuperWriter and the strong prompt workflow have the highest mean conflict grade. All methods have zero annotated OOC violations. Appendix~\ref{app:dramabench-diagnostics} defines the annotation units and scoring rules.

\subsection{Complete-script preference comparison}

The frozen human-preference model compares the selected complete scripts on the common fifteen-work cohort. Both presentation orders are evaluated; A/B/tie logits are aligned to candidate identities and averaged before applying softmax. The highest-probability class supplies the outcome. All ties in this comparison correspond to identical shared initial scripts.

\begin{table*}[t]
\centering
\begingroup
\setlength{\tabcolsep}{10pt}
\renewcommand{\arraystretch}{1.15}
\begin{tabularx}{\textwidth}{@{}Xcccc@{}}
\toprule
Baseline & Wins / Ties / Losses & Works & Preference (\%) & 95\% CI\\
\midrule
SuperWriter prompts & 10 / 3 / 2 & 15 & 76.7 & [57.1,93.3]\\
Dramaturge workflow & 11 / 3 / 1 & 15 & 83.3 & [66.7,96.4]\\
Strong prompt & 8 / 4 / 3 & 15 & 66.7 & [46.4,85.7]\\
\bottomrule
\end{tabularx}
\endgroup
\caption{Complete-script predictions of the frozen human-preference model. Wins, ties, and losses count predicted outcomes for SceneWeave; preference counts ties as half a win. Author-cluster intervals use 10,000 bootstrap samples. The same model selects candidates and predicts comparison outcomes.}
\label{tab:auxiliary-preferences}
\end{table*}

Table~\ref{tab:auxiliary-preferences} reports more predicted wins than losses for SceneWeave against all three baselines.

\subsection{Hybrid-writing case analysis}

We analyze four scenes to examine how particular writing choices affect dramatic treatment. Both workflows use the scene-first method and share the source, brief, initial plan, and draft. The hybrid uses the fine-tuned local writer (Appendix~\ref{app:writer}) for scene trials and \texttt{gpt-6-luna} for plan revision and complete-script integration; the API-only workflow uses \texttt{gpt-6-luna} throughout.

\paragraph{Care and resentment.}
In \emph{Lady Chatterley's Lover}, Clifford wins money while depending on Mrs. Bolton's company. The hybrid integrates a local trial's ``You're losing.'' / ``I know.'' with her continued dealing and private resentment. Reader comments value how the sparse exchange allows care, indulgence, and resistance to coexist. The source supplies the contradiction; local writing gives it an additional spoken form.

\paragraph{Impulse and restraint.}
In \emph{I Am Legend}, the API-only scene leaves Neville's hand still. The hybrid begins and interrupts a reach, giving his inward accusation a visible action to address. A local trial adds ``His thumb rubs its edge,'' joining the cross's experimental use with a possible self-soothing gesture. Reader comments recognize the hesitation while also valuing the API-only scene's spatial treatment of the earlier injection.

\paragraph{Private judgment and public authority.}
George delays his lesson while students talk. The hybrid voices his interpretation of Dreyer's support as an investment in future authority; a subsequent whisper unsettles it. The API-only scene conveys his need for recognition through a partial head-turn, faltering smile, and renewed audience contact. The difference concerns how private judgment becomes available to viewers through voice or disrupted public performance.

\paragraph{Contact and interpretation.}
In \emph{David Copperfield}, the hybrid prolongs a gesture from the shared initial draft: Mowcher questions David's prejudice while their hands remain joined. The local trial expands the pause and distance cue. Readers recognize its ambiguity, but prefer the API-only scene's clearer step, release, and level gaze. Their comments criticize the hybrid's explanatory directions: a meaningful gesture loses interpretive space when its significance is immediately stated.

\section{Conclusion}

SceneWeave writes alternative encounters and uses them to revise characterization and the surrounding adaptation. The four-reader study favors SceneWeave over the SuperWriter prompt workflow and both mechanism controls in aggregate; the qualitative cases identify specific uses of dialogue, action, and viewpoint that shape reader response.

\section*{Limitations}

The study covers prose passages and short connected scripts. The primary blind study includes four theatre graduate students and eight works. Translation and reader familiarity may influence preferences, while model-annotated content grades depend on annotation accuracy. Complete-script preference predictions use the same model for candidate selection and evaluation. Two scene alternatives cover only part of the adaptation space, and integration may lose useful details. The preference model uses human and synthetic supervision; incomplete source provenance and possible pretraining exposure limit unseen-work claims. Future work can investigate whether larger writing models trained on more high-quality adaptation and human-preference data improve script quality while maintaining alignment with reader preferences. Full-novel adaptation and broader audience reception remain to be studied.

\section*{Ethical considerations}

Generative artificial intelligence tools assisted with manuscript language polishing and literature search. The authors are responsible for the manuscript and the accuracy of its citations.

\bibliography{references}

\begin{thebibliography}{32}
\providecommand{\natexlab}[1]{#1}

\bibitem[{{Arena}()}]{arena-public-data}
{Arena}.
\newblock \href
  {https://huggingface.co/datasets/lmarena-ai/arena-human-preference-140k}
  {Public human preference data: 140k, 100k, and 55k releases}.
\newblock Official dataset repositories.
\newblock Additional releases: lmarena-ai/arena-human-preference-100k and
  lmarena-ai/arena-human-preference-55k.

\bibitem[{Artemova et~al.(2025)Artemova, Lucas, Venkatraman, Lee, Tilga,
  Uchendu, and Mikhailov}]{toloka-beemo}
Ekaterina Artemova, Jason~S Lucas, Saranya Venkatraman, Jooyoung Lee, Sergei
  Tilga, Adaku Uchendu, and Vladislav Mikhailov. 2025.
\newblock \href {https://doi.org/10.18653/v1/2025.naacl-long.357} {{Beemo}:
  Benchmark of expert-edited machine-generated outputs}.
\newblock In \emph{Proceedings of the 2025 Conference of the Nations of the
  Americas Chapter of the Association for Computational Linguistics: Human
  Language Technologies (Volume 1: Long Papers)}, pages 6992--7018. Association
  for Computational Linguistics.

\bibitem[{August and Gerwig(2020)}]{gerwig-august-2020-scriptnotes}
John August and Greta Gerwig. 2020.
\newblock \href
  {https://johnaugust.com/2020/scriptnotes-ep-433-the-one-with-greta-gerwig}
  {Scriptnotes, ep 433: The one with greta gerwig transcript}.
\newblock Scriptnotes interview transcript, January 16.

\bibitem[{Birch(2024)}]{birch-2024-lecture}
Alice Birch. 2024.
\newblock \href
  {https://www.bafta.org/media-centre/press-releases/screenwriters-lecture-series-2024-alice-birch/}
  {Screenwriters' lecture series 2024: Alice birch}.
\newblock BAFTA lecture transcript, published November 11.

\bibitem[{Chakrabarty et~al.(2025{\natexlab{a}})Chakrabarty, Laban, and
  Wu}]{chakrabarty-etal-2025-writing-rewards}
Tuhin Chakrabarty, Philippe Laban, and Chien-Sheng Wu. 2025{\natexlab{a}}.
\newblock \href {https://openreview.net/forum?id=jeDYcjuZIV} {{AI}-slop to
  {AI}-polish? aligning language models through edit-based writing rewards and
  test-time computation}.
\newblock In \emph{Second Conference on Language Modeling}.

\bibitem[{Chakrabarty et~al.(2025{\natexlab{b}})Chakrabarty, Laban, and
  Wu}]{chakrabarty-etal-2025-lamp}
Tuhin Chakrabarty, Philippe Laban, and Chien-Sheng Wu. 2025{\natexlab{b}}.
\newblock \href {https://doi.org/10.1145/3706598.3713559} {Can {AI} writing be
  salvaged? mitigating idiosyncrasies and improving human-{AI} alignment in the
  writing process through edits}.
\newblock In \emph{Proceedings of the 2025 CHI Conference on Human Factors in
  Computing Systems}, pages 1--33, New York, NY, USA. Association for Computing
  Machinery.

\bibitem[{Chen et~al.(2025)Chen, Yang, Lin, Mei, and
  Byrne}]{chen-etal-2025-dpo-ties}
Jinghong Chen, Guangyu Yang, Weizhe Lin, Jingbiao Mei, and Bill Byrne. 2025.
\newblock \href {https://arxiv.org/abs/2409.17431v2} {On extending direct
  preference optimization to accommodate ties}.
\newblock In \emph{Advances in Neural Information Processing Systems}.

\bibitem[{Fein et~al.(2026)Fein, Russo, Xiang, Jolly, Rafailov, and
  Haber}]{fein-etal-2026-litbench}
Daniel Fein, Sebastian Russo, Violet Xiang, Kabir Jolly, Rafael Rafailov, and
  Nick Haber. 2026.
\newblock \href {https://doi.org/10.18653/v1/2026.eacl-long.362} {{LitBench}: A
  benchmark and dataset for reliable evaluation of creative writing}.
\newblock In \emph{Proceedings of the 19th Conference of the European Chapter
  of the Association for Computational Linguistics (Volume 1: Long Papers)},
  pages 7740--7755, Rabat, Morocco. Association for Computational Linguistics.

\bibitem[{{Focus Features}(2024)}]{focus-2024-straughan}
{Focus Features}. 2024.
\newblock \href
  {https://www.focusfeatures.com/article/conclave_interview_peter_straughan}
  {Conclave's script weaves a web of intrigue and wonder}.
\newblock Interview with screenwriter Peter Straughan.

\bibitem[{Frick et~al.(2024)Frick, Li, Chen, Chiang, Angelopoulos, Jiao, Zhu,
  Gonzalez, and Stoica}]{frick-etal-2024-ppe}
Evan Frick, Tianle Li, Connor Chen, Wei-Lin Chiang, Anastasios~N. Angelopoulos,
  Jiantao Jiao, Banghua Zhu, Joseph~E. Gonzalez, and Ion Stoica. 2024.
\newblock \href {https://arxiv.org/abs/2410.14872} {How to evaluate reward
  models for {RLHF}}.
\newblock \emph{Preprint}, arXiv:2410.14872.

\bibitem[{Hu et~al.(2022)Hu, Shen, Wallis, Allen-Zhu, Li, Wang, Wang, and
  Chen}]{hu-etal-2022-lora}
Edward~J. Hu, Yelong Shen, Phillip Wallis, Zeyuan Allen-Zhu, Yuanzhi Li, Shean
  Wang, Lu~Wang, and Weizhu Chen. 2022.
\newblock \href {https://openreview.net/forum?id=nZeVKeeFYf9} {{LoRA}: Low-rank
  adaptation of large language models}.
\newblock In \emph{The Tenth International Conference on Learning
  Representations}.

\bibitem[{Huot et~al.(2025)Huot, Amplayo, Palomaki, Jakobovits, Clark, and
  Lapata}]{huot-etal-2025-agents-room}
Fantine Huot, Reinald~Kim Amplayo, Jennimaria Palomaki, Alice~Shoshana
  Jakobovits, Elizabeth Clark, and Mirella Lapata. 2025.
\newblock \href
  {https://proceedings.iclr.cc/paper_files/paper/2025/hash/0fbc8a83d93dd8021a4dd8d2d34138eb-Abstract-Conference.html}
  {Agents' room: Narrative generation through multi-step collaboration}.
\newblock In \emph{The Thirteenth International Conference on Learning
  Representations}.

\bibitem[{Lin et~al.(2025)Lin, Xiao, Mo, Zhang, Wang, Chen, Zhang, Zhang, Liu,
  Fang, and Xu}]{lin-etal-2025-reader-rewriter}
Zefeng Lin, Yi~Xiao, Zhiqiang Mo, Qifan Zhang, Jie Wang, Jiayang Chen, Jiajing
  Zhang, Hui Zhang, Zhengyi Liu, Xianyong Fang, and Xiaohua Xu. 2025.
\newblock \href {https://doi.org/10.48550/arXiv.2503.15655} {{$R^2$}: A {LLM}
  based novel-to-screenplay generation framework with causal plot graphs}.
\newblock \emph{Preprint}, arXiv:2503.15655.

\bibitem[{Ma et~al.(2025)Ma, Huang, and Lin}]{ma-etal-2025-dramabench}
Shijian Ma, Yunqi Huang, and Yan Lin. 2025.
\newblock \href {https://arxiv.org/abs/2512.19012} {{DramaBench}: A
  six-dimensional evaluation framework for drama script continuation}.
\newblock \emph{Preprint}, arXiv:2512.19012.

\bibitem[{Ma et~al.(2026)Ma, Huang, and Lin}]{ma-etal-2026-drama-failures}
Shijian Ma, Yunqi Huang, and Yan Lin. 2026.
\newblock \href {https://openreview.net/forum?id=9sVGZgUIFz} {I can't believe
  {LLMs} still can't write drama: Multi-dimensional failures in script
  continuation}.
\newblock In \emph{ICLR 2026 Workshop: I Can't Believe It's Not Better: Where
  Large Language Models Need to Improve}.

\bibitem[{Mirowski et~al.(2023)Mirowski, Mathewson, Pittman, and
  Evans}]{mirowski-etal-2023-dramatron}
Piotr Mirowski, Kory~W. Mathewson, Jaylen Pittman, and Richard Evans. 2023.
\newblock \href {https://doi.org/10.1145/3544548.3581225} {Co-writing
  screenplays and theatre scripts with language models: Evaluation by industry
  professionals}.
\newblock In \emph{Proceedings of the 2023 CHI Conference on Human Factors in
  Computing Systems}, pages 1--34, New York, NY, USA. Association for Computing
  Machinery.

\bibitem[{{Qwen Team}(2026)}]{qwen-2026-qwen35}
{Qwen Team}. 2026.
\newblock \href {https://huggingface.co/Qwen/Qwen3.5-9B} {{Qwen3.5-9B}:
  Official model card}.
\newblock Hugging Face model repository.

\bibitem[{Rafailov et~al.(2023)Rafailov, Sharma, Mitchell, Manning, Ermon, and
  Finn}]{rafailov-etal-2023-dpo}
Rafael Rafailov, Archit Sharma, Eric Mitchell, Christopher~D. Manning, Stefano
  Ermon, and Chelsea Finn. 2023.
\newblock \href
  {https://proceedings.neurips.cc/paper_files/paper/2023/hash/a85b405ed65c6477a4fe8302b5e06ce7-Abstract-Conference.html}
  {Direct preference optimization: Your language model is secretly a reward
  model}.
\newblock In \emph{Advances in Neural Information Processing Systems},
  volume~36.

\bibitem[{{roytian1992}()}]{stage-data}
{roytian1992}.
\newblock \href {https://github.com/roytian1992/STAGE_v0} {{STAGE}: Evolving
  stories}.
\newblock GitHub repository.

\bibitem[{Sciamma(2019)}]{sciamma-2019-lecture}
C{\'e}line Sciamma. 2019.
\newblock \href
  {https://www.bafta.org/media-centre/press-releases/screenwriters-lecture-series-2019-celine-sciamma-2/}
  {Screenwriters' lecture series 2019: C{\'e}line sciamma}.
\newblock BAFTA lecture transcript.

\bibitem[{{Strands Agents}(2026)}]{strands-2026-decider}
{Strands Agents}. 2026.
\newblock \href
  {https://huggingface.co/StrandsAgents/strands-decider-2B-hobson-v19}
  {{strands-decider-2B-hobson-v19}: Official model card}.
\newblock Hugging Face model repository.

\bibitem[{{supergeniodelmale}()}]{singlescenes-data}
{supergeniodelmale}.
\newblock \href {https://github.com/supergeniodelmale/SingleScenes}
  {{SingleScenes}: A simple benchmark for {LLM} film script understanding and
  generation}.
\newblock GitHub repository.

\bibitem[{Tang et~al.(2026)Tang, Chen, Li, Xie, Ma, and
  Qu}]{tang-etal-2026-duodrama}
Yuying Tang, Xinyi Chen, Haotian Li, Xing Xie, Xiaojuan Ma, and Huamin Qu.
  2026.
\newblock \href {https://doi.org/10.1145/3772318.3790568} {{DuoDrama}:
  Supporting screenplay refinement through {LLM}-assisted human reflection}.
\newblock In \emph{Proceedings of the 2026 CHI Conference on Human Factors in
  Computing Systems}, New York, NY, USA. Association for Computing Machinery.

\bibitem[{Wang et~al.(2024{\natexlab{a}})Wang, Li, Chen, Cai, Zhu, Lin, Cao,
  Kong, Liu, Liu, and Sui}]{wang-etal-2024-fair-evaluators}
Peiyi Wang, Lei Li, Liang Chen, Zefan Cai, Dawei Zhu, Binghuai Lin, Yunbo Cao,
  Lingpeng Kong, Qi~Liu, Tianyu Liu, and Zhifang Sui. 2024{\natexlab{a}}.
\newblock \href {https://doi.org/10.18653/v1/2024.acl-long.511} {Large language
  models are not fair evaluators}.
\newblock In \emph{Proceedings of the 62nd Annual Meeting of the Association
  for Computational Linguistics (Volume 1: Long Papers)}, pages 9440--9450,
  Bangkok, Thailand. Association for Computational Linguistics.

\bibitem[{Wang et~al.(2024{\natexlab{b}})Wang, Bukharin, Delalleau, Egert,
  Shen, Zeng, Kuchaiev, and Dong}]{wang-etal-2024-helpsteer2-preference}
Zhilin Wang, Alexander Bukharin, Olivier Delalleau, Daniel Egert, Gerald Shen,
  Jiaqi Zeng, Oleksii Kuchaiev, and Yi~Dong. 2024{\natexlab{b}}.
\newblock \href {https://arxiv.org/abs/2410.01257} {{HelpSteer2-Preference}:
  Complementing ratings with preferences}.
\newblock \emph{Preprint}, arXiv:2410.01257.

\bibitem[{Wang et~al.(2025{\natexlab{a}})Wang, Zeng, Delalleau, Egert, Evans,
  Shin, Soares, Dong, and Kuchaiev}]{wang-etal-2025-helpsteer3-edit}
Zhilin Wang, Jiaqi Zeng, Olivier Delalleau, Daniel Egert, Ellie Evans,
  Hoo-Chang Shin, Felipe Soares, Yi~Dong, and Oleksii Kuchaiev.
  2025{\natexlab{a}}.
\newblock \href {https://arxiv.org/abs/2503.04378} {{HelpSteer3}:
  Human-annotated feedback and edit data to empower inference-time scaling in
  open-ended general-domain tasks}.
\newblock \emph{Preprint}, arXiv:2503.04378.

\bibitem[{Wang et~al.(2025{\natexlab{b}})Wang, Zeng, Delalleau, Shin, Soares,
  Bukharin, Evans, Dong, and Kuchaiev}]{wang-etal-2025-helpsteer3-preference}
Zhilin Wang, Jiaqi Zeng, Olivier Delalleau, Hoo-Chang Shin, Felipe Soares,
  Alexander Bukharin, Ellie Evans, Yi~Dong, and Oleksii Kuchaiev.
  2025{\natexlab{b}}.
\newblock \href {https://arxiv.org/abs/2505.11475} {{HelpSteer3-Preference}:
  Open human-annotated preference data across diverse tasks and languages}.
\newblock \emph{Preprint}, arXiv:2505.11475.

\bibitem[{Wu et~al.(2026)Wu, Bai, Hu, Li, and Lee}]{wu-etal-2026-superwriter}
Yuhao Wu, Yushi Bai, Zhiqiang Hu, Juanzi Li, and Roy Ka-Wei Lee. 2026.
\newblock \href {https://doi.org/10.18653/v1/2026.findings-acl.428}
  {{SuperWriter}: Reflection-driven long-form generation with large language
  models}.
\newblock In \emph{Findings of the Association for Computational Linguistics:
  ACL 2026}, pages 8790--8812, San Diego, California, United States.
  Association for Computational Linguistics.

\bibitem[{Xie et~al.(2025)Xie, Guo, Jing, Wang, Lv, and
  Wang}]{xie-etal-2025-dramaturge}
Wenda Xie, Chao Guo, Yanqing Jing, Junle Wang, Yisheng Lv, and Fei-Yue Wang.
  2025.
\newblock \href {https://doi.org/10.48550/arXiv.2510.05188} {Plug-and-play
  dramaturge: A divide-and-conquer approach for iterative narrative script
  refinement via collaborative {LLM} agents}.
\newblock \emph{Preprint}, arXiv:2510.05188.
\newblock Version 4, revised April 2026.

\bibitem[{Yang et~al.(2023)Yang, Klein, Peng, and Tian}]{yang-etal-2023-doc}
Kevin Yang, Dan Klein, Nanyun Peng, and Yuandong Tian. 2023.
\newblock \href {https://doi.org/10.18653/v1/2023.acl-long.190} {{DOC}:
  Improving long story coherence with detailed outline control}.
\newblock In \emph{Proceedings of the 61st Annual Meeting of the Association
  for Computational Linguistics (Volume 1: Long Papers)}, pages 3378--3465,
  Toronto, Canada. Association for Computational Linguistics.

\bibitem[{Yang et~al.(2022)Yang, Tian, Peng, and Klein}]{yang-etal-2022-re3}
Kevin Yang, Yuandong Tian, Nanyun Peng, and Dan Klein. 2022.
\newblock \href {https://doi.org/10.18653/v1/2022.emnlp-main.296} {{Re3}:
  Generating longer stories with recursive reprompting and revision}.
\newblock In \emph{Proceedings of the 2022 Conference on Empirical Methods in
  Natural Language Processing}, pages 4393--4479, Abu Dhabi, United Arab
  Emirates. Association for Computational Linguistics.

\bibitem[{Zheng et~al.(2025)Zheng, Song, Zhou, You, Zhan, Ma, Song, Lim, Chen,
  and Yang}]{cml-bench-data}
Mingzhe Zheng, Dingjie Song, Guanyu Zhou, Jun You, Jiahao Zhan, Xuran Ma,
  Xinyuan Song, Ser-Nam Lim, Qifeng Chen, and Harry Yang. 2025.
\newblock \href {https://arxiv.org/abs/2510.06231} {{CML-Bench}: A framework
  for evaluating and enhancing {LLM}-powered movie scripts generation}.
\newblock \emph{Preprint}, arXiv:2510.06231.

\end{thebibliography}

\appendix
\section{Implementation and Training Conditions}
\label{app:implementation}

\paragraph{Model inputs and outputs.}
Scene writing returns a scene identifier and text; complete writing returns a full script. Planning, interpretation, material assignment, and review use structured outputs. Matched conditions share the source, brief, initial plan and draft, and model assignments. Scenes with assigned trial material are rewritten in the revised context; other scenes are written from their plan descriptions.

\paragraph{Scene edits and revision selection.}
The complete set of proposed edits is validated and applied to the specified plan. A merge operation specifies its input scene identifiers and replaces the corresponding scene entries with the merged scene; a split replaces its parent with ordered new scenes. For each alternative edit, scenes for trial writing are selected after the edits are applied, including surviving replacements, insertions, moves, merges, and splits, in the resulting presentation order. Deletions redirect writing to surviving scenes adjacent to the deletion and to declared dependent scenes. Changes confined to overall adaptation choices retain the surviving focal scene as the default. The resulting scene list is checked before it is passed to the writing model.

The previous draft is retained only if review identifies unintended repetition in the revision and quotes a passage of at least 160 characters that occurs at least twice, with more occurrences than in the previous draft.

\paragraph{Public comparison sources.}
SuperWriter prompts are taken from official repository commit \texttt{436149d87908a462b5f74}\allowbreak\texttt{270cb43d797163f6e17}. Planning supplies the common outline; writing and refinement use the shared LLM. The complete instructions and generated screenplay text are preserved when adapting output fields.

Dramaturge permits at most two outer iterations, an improvement threshold of one on its original 100-point internal scale, and three consecutive unsuccessful detail iterations before stopping.

\paragraph{Preference model.}
We fine-tune a pretrained decision model with LoRA, retaining its pointer head for A, B, and tie. Human and synthetic comparisons train the model; human validation loss selects the checkpoint. Development preferences support calibration, with test labels excluded from model selection. The supplementary materials provide model revisions and training configurations.

\paragraph{Preference probability evaluation.}
For probability evaluation, A/B/tie logits from both presentation orders are aligned to the original candidate identities, averaged, and converted to probabilities with softmax. Disagreement between the two original judgments is reported separately.
\section{Data Sources and Preparation}
\label{app:data}

\paragraph{Public preference annotations.}
Arena releases provide user comparisons; PPE provides public pairwise judgments; HelpSteer2/3 provide human preferences, including individual signed judgments where available. Screening checks the requested task, both texts, available context, and dramatic form. We retain source identifiers, upstream partitions, annotation origin, and the study partition. Annotator explanations and response-model identities are excluded from comparison inputs.

\paragraph{Public writing supervision.}
LAMP supplies original drafts and professional revisions \citep{chakrabarty-etal-2025-lamp,chakrabarty-etal-2025-writing-rewards}. The Benchmark of Expert-edited Machine-generated Outputs (Beemo) supplies model drafts and human edits \citep{toloka-beemo}; its generation and rewriting records are filtered for narrative tasks. HelpSteer3's edit subset supplies conversational context, feedback, and a human-edited response \citep{wang-etal-2025-helpsteer3-edit}. These form draft-to-revision and feedback-conditioned revision examples. LAMP edit spans are aligned to the complete revised target and remain annotations within each example.

DramaBench supplies context--continuation pairs \citep{ma-etal-2025-dramabench}. We verify that the prefix and continuation reconstruct the released script. The Cinematic Markup Language Benchmark (CML-Bench) supplies screenplay segments \citep{cml-bench-data}; Extensible Markup Language (XML) fields are rendered in order, preserving scene descriptions, characters, dialogue, and parentheticals. Malformed structures are excluded. SingleScenes supplies scene synopses and scripts \citep{singlescenes-data}; the retained material comes from \emph{Eternal Sunshine of the Spotless Mind}. Synopsis-to-scene supervision uses these reference-derived synopses as inputs, with all examples from the film assigned to a single study partition.

\paragraph{Private texts and synthetic supervision.}
Private prose--adaptation pairs provide passage-to-scene examples, and a private screenplay collection provides adjacent-scene continuation examples. Public metadata from the bilingual narrative benchmark \emph{STAGE: Evolving Stories} supplies identifiers for part of the latter collection \citep{stage-data}. Passage correspondence is checked separately from a work-level adaptation link, using edition, source boundaries, scene boundaries, participants, and depicted events. The prose-only input omits its reference scene; revision tasks provide the existing scene as the draft input. Private source texts are not redistributed.

Synthetic examples retain their teacher identity, source group, task, and input--output record. Writing targets generated or rewritten by a model are labeled synthetic. Synthetic preference pairs use the same request for both drafts and preserve their annotation provenance. Filtering checks dramatic form, consistency with the brief, presentation-order agreement where applicable, and whether judgments reward fidelity or departure without a stated dramatic benefit.

\paragraph{Data partitioning and duplicate removal.}
Known work, adaptation, author, prompt, and revision-family links determine which examples belong to the same study group. Exact normalized-text duplicates and reversed copies of the same preference pair are removed. Writer records are screened against known held-out groups and text fingerprints; overlap review is triggered when at least 85\% of the distinct five-token sequences in the shorter text occur in the longer text, provided that at least 25 distinct sequences are available. Full inputs and targets are checked for missing fields, task compatibility, and compliance with token-length limits. Data manifests retain source versions, extraction rules, partitions, and original human labels.

\section{A Task-conditioned Local Writer}
\label{app:writer}

The local writer is obtained by fine-tuning Qwen3.5-9B with LoRA \citep{qwen-2026-qwen35,hu-etal-2022-lora} on adaptation, scene writing, continuation, and revision examples. Supervised learning uses complete target texts. Human preference alignment uses direct preference optimization with a Davidson extension for ties \citep{rafailov-etal-2023-dpo,chen-etal-2025-dpo-ties}, followed by task-specific scene training with supervised replay. Checkpoints are selected using fixed validation data.

The controlled method comparison uses the API model throughout. In the hybrid case study, the local writer generates scene trials and the API model revises the plan and integrates the complete script. Training configurations and model identifiers are supplied in the supplementary materials.

\section{Full-Script Content Diagnostics}
\label{app:dramabench-diagnostics}

Four human-labeled preference examples from the training set and their released explanations guide the annotation instructions. Their original pairwise choices are retained as context. Conflict annotations distinguish principal, supporting, and background tensions, with each criterion assessed within the requested adaptation scope. Direct expression, subtext, quiet encounters, confrontation, and open or resolved endings share the same grade anchors.

Complete drafts from the last finished iteration remain eligible after an interrupted later iteration; partial drafts are excluded. The first complete annotation under the fixed rubric is retained; scoring follows candidate selection.

A fixed unit is an action paragraph or a complete dialogue turn within a scene. Speaker cues on separate lines or before a colon identify dialogue. Headings mark scene boundaries; titles and transitions remain contextual material outside unit denominators. Coverage of eligible units is checked programmatically. The same fifteen source works enter every column and every baseline contrast.

Let $R,N$ count units labeled redundant and nonredundant, respectively; let $O,I,U$ count assessable dialogue turns labeled out-of-character, in-character, and neutral, respectively. Let $S,G$ count dialogue turns with specific and generic voices, respectively, and let $V,M$ count violated and maintained world constraints, respectively. The rates are
\begin{equation}
 \begin{aligned}
 r_{\mathrm{rep}}&=\frac{R}{R+N}, &
 r_{\mathrm{ooc}}&=\frac{O}{O+I+U},\\
 r_{\mathrm{voice}}&=\frac{S}{S+G}, &
 r_{\mathrm{logic}}&=\frac{V}{V+M}.
\end{aligned}
\end{equation}
Uncertain labels and irrelevant facts stay outside their respective denominators. Observed conduct and object states are world facts; unverified character reports remain separate. Zero error rates denote no annotated violations. Column means average per-script rates and are displayed as percentages.

For each dimension $d$ (emotion or conflict), reading $r$ assigns anchored levels $a_{d1r},a_{d2r},a_{d3r}\in\{0,1,2,3\}$ to the three criteria. The per-script grade is
\begin{equation}
 g_d(s)=\frac12\sum_{r=1}^{2}\operatorname{median}(a_{d1r},a_{d2r},a_{d3r}).
\end{equation}
For each method, the mean grade averages $g_d(s)$ over the fifteen works. Raw criterion levels and repeated readings are retained. Inapplicable or unassessable criteria are excluded.

Psychological particularity concerns these people's specific relationships and lived circumstances. Realization concerns how their experience becomes perceptible through particular expression and response, including effective direct speech. Development concerns the psychological course or sustained contradiction within the brief, without requiring an altered endpoint or more emotional events. Conflict tension concerns enacted incompatible needs or positions; development concerns motivated responses and consequences; integration concerns preparation, connections between scenes, and final treatment. Either dimension can receive the highest grade for a concentrated, simple, quiet, or comic treatment. Complexity, escalation, closure and ambiguity earn no automatic credit.

\section{Human Comparison Instructions}
\label{app:human}

Each pair includes a factual source card and brief, anonymous complete scenes A and B, and relevant surrounding text in script order. Omissions outside complete scenes are marked. Source cards do not specify character motivations or intended emotional effects. Source situation anchors, scene boundaries, and word counts are fixed before annotation. Scenes are matched by character roles and situations under permitted changes; cases without reliable correspondence are marked as unassessable.

Readers start with one pair and may continue with pairs from other works. Candidate order is randomized independently of method identity. Assignment favors pairs with fewer completed judgments and balances the three contrasts. Anonymous identifiers prevent repeated submission for the same pair.

Ties and inability to judge are recorded separately, with exact texts, context, and presentation order retained.

\end{document}